\documentclass{article}

\PassOptionsToPackage{numbers, compress}{natbib}
\usepackage[preprint]{neurips_2023}

\usepackage{amsmath,amsfonts,bm}

\def\eqref#1{equation~\ref{#1}}
\def\1{\bm{1}}

\def\vg{{\bm{g}}}

\def\vi{{\bm{i}}}

\def\vk{{\bm{k}}}

\def\vq{{\bm{q}}}

\def\vx{{\bm{x}}}

\DeclareMathAlphabet{\mathsfit}{\encodingdefault}{\sfdefault}{m}{sl}
\SetMathAlphabet{\mathsfit}{bold}{\encodingdefault}{\sfdefault}{bx}{n}

\usepackage[utf8]{inputenc} % allow utf-8 input
\usepackage[T1]{fontenc}    % use 8-bit T1 fonts
\usepackage{hyperref}       % hyperlinks
\usepackage{url}            % simple URL typesetting
\usepackage{booktabs}       % professional-quality tables
\usepackage{amsfonts}       % blackboard math symbols
\usepackage{amsmath}
\usepackage{amssymb}
\usepackage{nicefrac}       % compact symbols for 1/2, etc.
\usepackage{microtype}      % microtypography
\usepackage{xcolor}         % colors
\usepackage{graphicx}
\usepackage{wrapfig}
\usepackage{textgreek}

\DeclareSymbolFont{extraup}{U}{zavm}{m}{n}
\DeclareMathSymbol{\varheart}{\mathalpha}{extraup}{86}
\DeclareMathSymbol{\vardiamond}{\mathalpha}{extraup}{87}

\title{Retrieval Head Mechanistically Explains Long-Context Factuality}

\author{
  Wenhao Wu$^\text{\textlambda}$\quad\quad 
  Yizhong Wang$^\text{\textdelta}$\quad\quad 
  Guangxuan Xiao$^\text{\textsigma}$\quad\quad
  Hao Peng$^\text{\textpi}$\quad\quad
  Yao Fu$^\text{\textmu}$\\
  $^\text{\textlambda}$Peking University\quad 
  $^\text{\textdelta}$University of Washington\quad
  $^\text{\textsigma}$MIT\quad 
  $^\text{\textpi}$UIUC\quad $^\text{\textmu}$University of Edinburgh\\
  \texttt{waynewu@pku.edu.cn}\quad\quad 
  \texttt{haopeng@illinois.edu}\quad\quad 
  \texttt{yao.fu@ed.ac.uk} \\
  \url{https://github.com/nightdessert/Retrieval_Head}\\ 
}

\begin{document}

\maketitle
\begin{abstract}
Despite the recent progress in long-context large language models (LLMs), it remains elusive how these transformer-based language models acquire the capability to retrieve relevant information from arbitrary locations within the long context. This paper aims to address this question.
Our systematic investigation across 4 model families, 6 model scales, and 3 types of finetuning reveals that a special type of attention heads are largely responsible for retrieving relevant information from long context, which we dub \emph{retrieval heads}.
We identify important and intriguing properties of retrieval heads:
(1) \textit{universal}: 
 all the explored models with long-context capability have a set of retrieval heads;
(2) \textit{sparse}: only a small portion (less than 5\%) of the attention heads are retrieval. 
(3) \textit{intrinsic}: retrieval heads already exist in
models pretrained with short context.
When extending the context length to 32-128K by continual pretraining, 
it is still the same set of heads that perform information retrieval.
(4) \textit{dynamically activated}: 
take Llama-2 7B for example, 12 
retrieval heads always attend to the required information no matter how the context is changed.
The rest of the retrieval heads are activated in different contexts.
(5) \textit{causal}: 
completely pruning retrieval heads leads to failure in retrieving relevant information and results in hallucination, while pruning random non-retrieval heads does not affect the model's retrieval ability.
We further show that retrieval heads strongly influence 
chain-of-thought (CoT) reasoning, where the model needs to frequently refer back the question and previously-generated context.
Conversely, tasks where the model directly generates the answer using its intrinsic knowledge
are less impacted by masking out retrieval heads. 
These observations collectively explain which internal part of the model seeks information from the input tokens.
We believe our insights on retrieval heads foster future research on reducing hallucination, improving reasoning, and compressing the KV cache.
\end{abstract}

\section{Introduction}
\label{sec:intro}
\begin{figure*}[!t]
\small
  \centering
  \includegraphics[width=\linewidth]{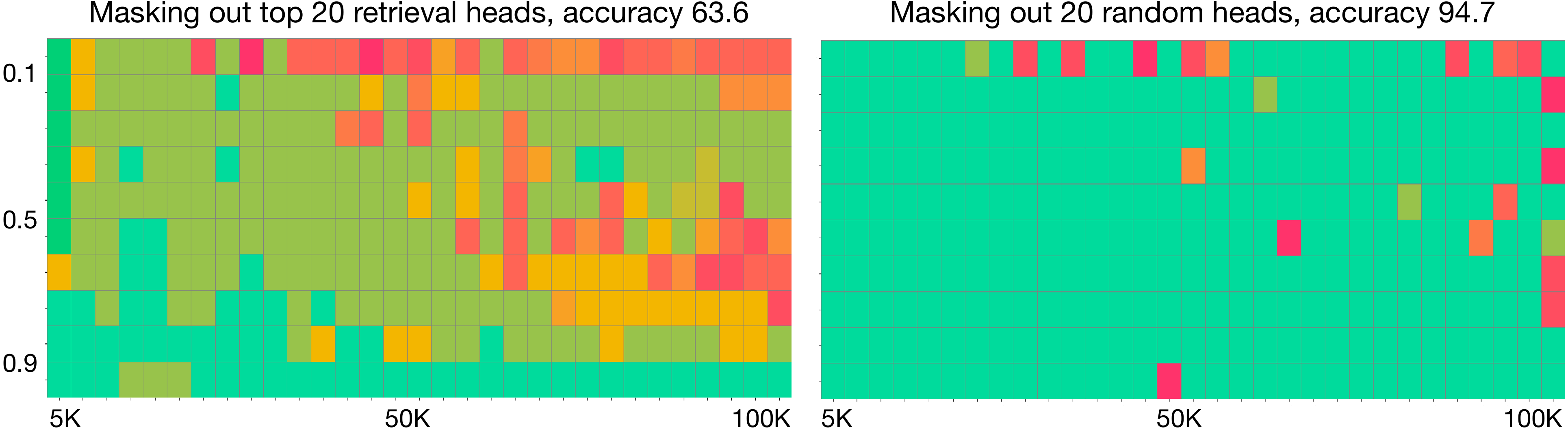}
  \caption{Retrieval heads are the heads that redirect information from the input to the output. Left: masking out the top retrieval heads of LLaMA 2 7B 80K, its Needle-in-a-Haystack performance drops significantly, and the model hallucinates during decoding. Right: masking out random non-retrieval heads does not influence the model's needle-in-a-haystack behavior. 
  We further note that the retrieval head only influence factuality but not the language capability: when they are masked, the model hallucinates by saying ``go to the beach'', which is still a fluent sentence, but not factual (as the correct answer is ``eating a sandwich at Dolores park.'').
  }
  \label{fig:retrieval_head}
\end{figure*}

This work studies the internal mechanism of how long-context language models can utilize information at arbitrary locations within the input. 
Recent advances in long-context language modeling~\citep{claude, reid2024gemini, fu2024data} show inspiring results, particularly on the Needle-in-a-Haystack test~\citep{needleinhaystack}, which asks the model to precisely retrieve the information of a short sentence (the needle) within a long context (the haystack). 
Such capability is the basis of more advanced long-context tasks, which usually interleaves retrieval and reasoning in a multi-step fashion~\citep{kuratov2024search}. 
Based on extensive experiments across 4 model families, 6 model scales, and 3  types of finetuning,
we show that across the models' attention layers, there exist a small number of retrieval heads that search the information being asked, and redirect the relevant tokens from the input to the output. 
Activation of retrieval heads explains whether the output is factual or hallucinated. 
When such heads are activated, the model behaves faithful to the input document.
When they are not activated, or intentionally masked out in controlled experiments (Fig.~\ref{fig:retrieval_head}), the model cannot find the relevant information and hallucinate instead.

% P2. Motivate retrieval head with what attention might be doing, the copy net, and induction head
The discovery of the retrieval head is motivated by the question of what the attention mechanism is doing when the model can or cannot find the given needle.
We take important inspiration from two existing works: the CopyNet~\citep{gu-etal-2016-incorporating} and the Induction Head~\citep{olsson2022context}.
The CopyNet is essentially a single-layer, single-head attention mechanism in the age of RNNs that copy-paste tokens from the input to the output.
Induction Heads~\citep{olsson2022context} are special heads within a multi-layer, multi-head attention network that implements an implicit program induction algorithm.
Combining the observation from the two works, we natually hypothesize that, just like induction heads are accountable for in-context learning, there might exist special heads that are accountable for information retrieval and implement a conditional copy-paste algorithm.

We design algorithms to detect retrieval heads within the transformer architecture (Sec.~\ref{sec:retrieval_head}), and conduct large-scale experiments to demonstrate important properties of them (Sec.~\ref{sec:properties}):
(1) retrieval heads are universal and sparse: for any model family (LLaMA~\citep{touvron2023llama}, Yi~\citep{young2024yi}, QWen~\citep{bai2023qwen} and Mistral~\citep{jiang2023mistral}), at any scale (6B, 14B, and 34B and 8$\times$7B), either base or chat, either dense or MoE, as long as the model can precisely recite the input information, they have a small number of retrieval heads (Fig.~\ref{fig:retrieval_head});
(2) they are intrinsic: the base model (e.g., LLaMA2 base) already contains retrieval heads (as a consequence of large-scale pretraining). Subsequent derivations, such as the long-context continue pretraining (LLaMA2 7B 80K),  chat fine-tuning (Qwen Chat), or even sparse upcycling~\citep{komatsuzaki2022sparse, jiang2024mixtral} uses the same retrieval heads as the base model (Fig.~\ref{fig:heat_map}); 
(3) they are dynamically activated according to the context: the strongest retrieval heads (e.g., 13 for LLaMA 2 7B) are always activated no matter what the required information is, while weaker retrieval heads  are activated on different parts of the required information; consequently these heads compensate each other's functionality: removing a subset of the heads, the model at least retrieve part of the required information;
(4) the retrieval heads are causal: 
say we put a needle "the best thing to do in San Francisco is to eat a sandwich in Dolores Park on a sunny day",
completely masking out retrieval heads, the model hallucinates (by saying the best thing is to visit Golden Gate bridge); 
partially masking out the heads, the model retrieves part of the needle (e.g., it gets the sandwich but forget the Dolores Park);
masking out random non-retrieval heads, the model still find full needle; 
when we do not mask the head yet the model still hallucinate in some cases, 
the retrieval heads are not activated. 
We further note that chain-of-thought reasoning also heavily relies on retrieval heads because the model needs to refer back the input information, indicating a complex relationship between the model's retrieval and reasoning capability.

The discovery of retrieval head has profound implications on long-context modeling:
(1) it marks a significant step forward in the field of mechanistic interpretability~\citep{bricken2023monosemanticity, olsson2022context} because for the first time we pin point a particular subnet implementing the conditional retrieval algorithm;
(2) it explains why certain context-compression methods fail to keep factuality (because they removes the retrieval head, e.g., in~\citealt{xiao2023efficient}), and suggests future research on KV cache compression~\citep{ge2023model, kang2024gear}, a key problem for deploying long-context models, should consider the influence of retrieval heads.

% \section{Preliminary}
% \label{sec:preliminary}
% \input{010_preliminary}

\section{Detecting Retrieval Head}
\label{sec:retrieval_head}
\begin{figure*}[!t]
\small
  \centering
  \includegraphics[width=\linewidth]{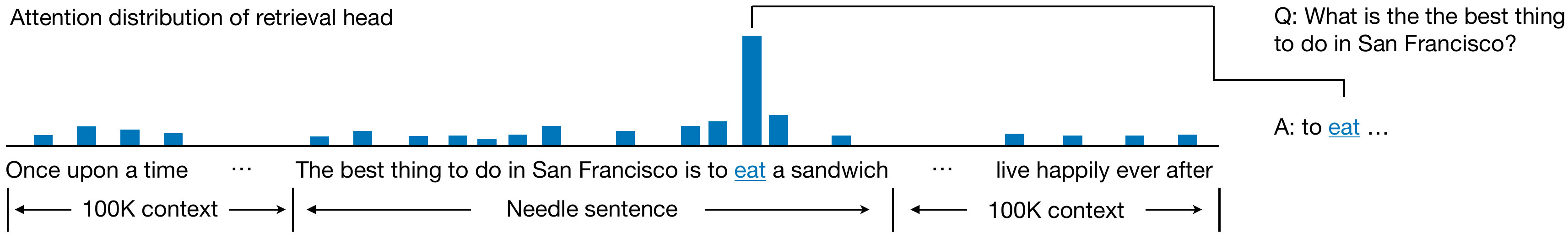}
  \caption{An attention head that performs a copy-paste operation if the token that the head attend to is the same the token being generated. 
  The retrieval score of a head is defined as the frequency of this head's copy-paste behavior when answering quesions that asks for raw information from the input. 
  }
  \label{fig:retrieval_attention_dist}
\end{figure*}

\begin{table*}[t!]
    \centering
    \caption{\label{tab:all_models} We consider a wide range of language model families and show that the basic properties of retrieval heads are universal and consistent across all language models we study. }
    \begin{tabular}{@{}l|l|l@{}}
    \toprule
     \bf  Base Model& \bf Variant & \bf Variation Type\\
         \midrule

     Llama-2-7B& Llama-2-7B-80K & Length Extension via Continue Pretrain\\
     &Llama-2-13B-64K & Model Scaling and Length Extension\\
          \midrule

     Mistral-7B-v0.2 & Mistral-7B-Instruct-v0.2 & SFT and RLHF\\
     & Mixtral-8x7B-v0.1 & Sparse Upcycling to Mixture of Experts\\
     \midrule
     Yi-6B      & Yi-6B-200K & Length Extension via Continue Pretrain\\
                & Yi-34B-200K & Model Scaling and Length Extension\\
     \midrule
     Qwen1.5-14B & Qwen1.5-14B-Chat& SFT and RLHF\\
     \toprule
    \end{tabular}
\end{table*}

To detect which head is implementing the retrieval algorithm, we introduct a \textit{retrieval score} to measures the frequency of a head's copy-paste behavior during autoregressive decoding. 
An attention head with high retrieval score suggests that statistically across various contexts, this head is frequently copying the input tokens from the input to the output.

\textbf{Needle-in-a-Haystack}\quad\quad Our retrieval head detection algorithm roots from the needle-in-a-Haystack test, which asks the model to copy-paste the input tokens to the output.
Given a question $\vq$ and its corresponding answer $\vk$ (the needle),  we insert $\vk$ in a given context  $\vx$  (the haystack) at a random position index range $\vi_q$.
The language model is then tasked with answering $\vq$ based on the haystack with the inserted needle.
We set $\vq$ and $\vk$ unique and irrelevant with the given long context,
ensuring that if an answer is correctly generated, it is indeed copied from the context, not from the model's internal knowledge.

\textbf{Retrieval Score for Attention Heads}\quad\quad 
We define the retrieval score as the frequency of a head's copy-paste operations. 
Specifically, 
during auto-regressive decoding (we use greedy decoding by default), 
denote the current token being generated as $w$ and
the attention scores of a head as $\boldsymbol{a} \in \mathcal{R}^{|\vx|}$.
As demonstrated in Fig.~\ref{fig:retrieval_attention_dist}, we say an attention head $h$ copies and pastes a token from the needle to the output sentence if it follows two criteria: 
(1) $w \in \vk$, i.e., $w$ is a token within the needle sentence. (2)
$\vx_j= w, j = \arg\max(\boldsymbol{a}), j \in \vi_q$, i.e., the input token that receives the most attention probability mass by this head is a token within the needle and is the same token as the currently generated token. 
Let $\vg_h$ be the set containing all tokens copy and pasted by a given head $h$, we define:
\begin{equation}
    \text{Retrieval score for head}\;\; h = \frac{|\vg_h \cap \vk |}{|\vk|}, 
\end{equation}
Intuitively, retrieval score  represents a token-level recall rate of the most attended tokens by an attention head.
For example, when retrieving a needle of 10 tokens, a retrieval score of 0.9 indicates that the attention head has copies and pasted 9 tokens in the 10-token target answer.

\begin{figure*}[!t]
\small
  \centering
  \includegraphics[width=\linewidth]{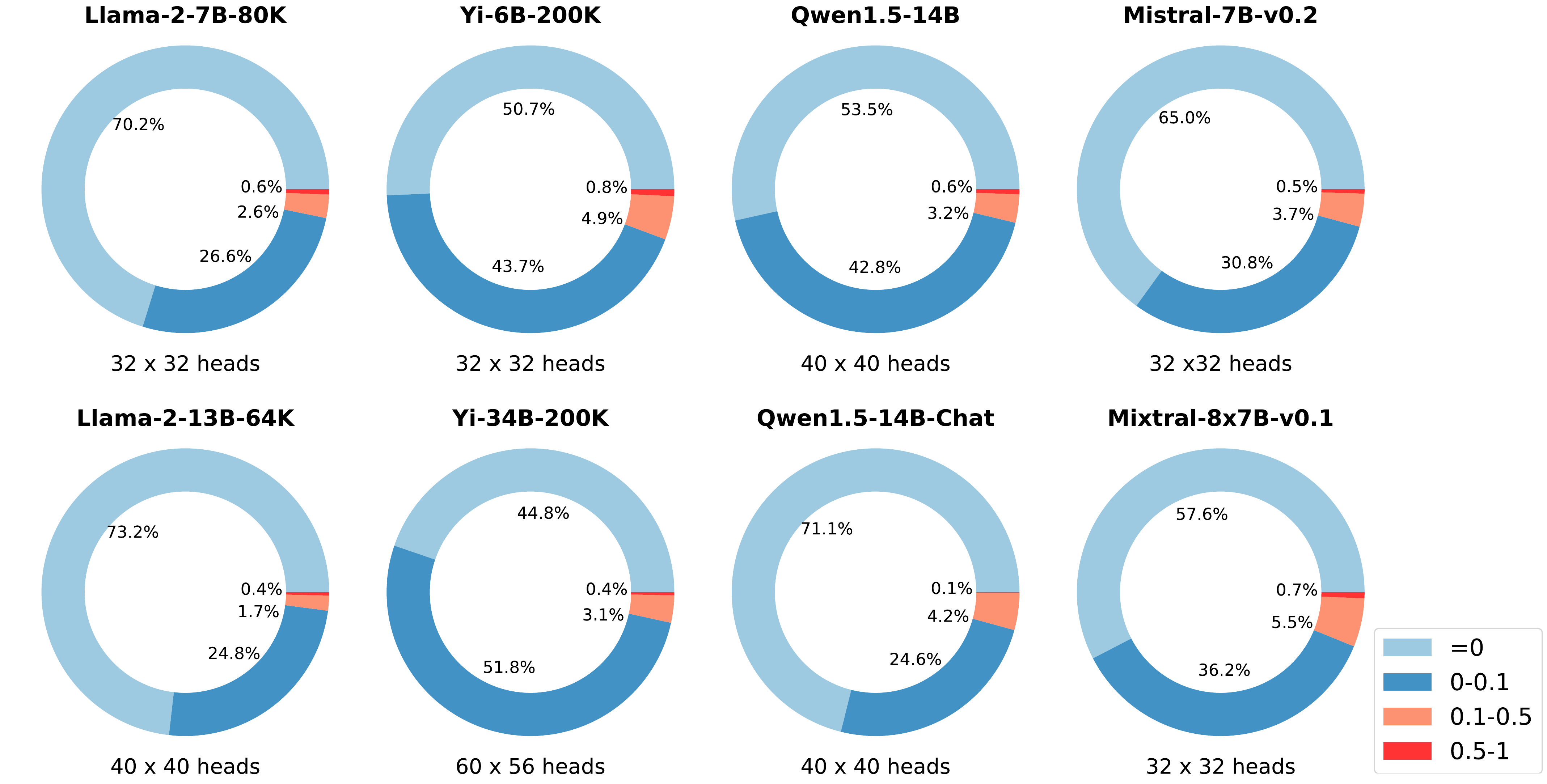}
  \caption{Retrieval heads are universal and sparse across model family and scale. 
  For all models we consider, less than 5\% of the attention heads are activated more than 50\% of the time (with a retrieval score higher than 0.5) when retrieval is required.
  }
  \label{fig:score_range}
\end{figure*}

\textbf{Retrieval Head Detection Algorithm}\quad\quad
We calculate the retrieval score for all attention heads under a diverse set of input contexts. 
For each language model we consider, we compile three sets of Needle-in-a-Haystack samples, each consisting of a unique tuple $(\vq, \vk, \vx)$.
For each sample, we make sure $(\vq, \vk)$ is semantically irrelevant with $\vx$ and that $\vq$ cannot be answered using the model's existing knowledge by manually inspecting the model output. 
Then for each $(\vq, \vk, \vx)$ sample, we perform  Needle-in-a-Haystack on 20 different length values uniformly sampled from 1K-50K, where in each length, $\vq$ is inserted in 10 different depth uniformly ranging from the start to the end of $\vx$.
We note that this scale of tests gives stable outputs as the average retrieval score converges after just a few samples. 
In total, each language model is subjected to approximately 600 instances of retrieval testing. 
We calculate the retrieval score for each attention head in each test and use the average of these scores as the head's final retrieval score.
The attention heads with relatively larger retrieve score  can be considered as retrieval head.
In our case (Fig.~\ref{fig:score_range}), we set the threshold as 0.1, meaning that as long as the head performs copy-paste in 10\% of the times, we consider it a retrieval head.

\section{Basic Properties of Retrieval Heads}
\label{sec:properties}

This section discusses important properties of retrieval heads:
(1) universal and sparse: any model that exhibits long-context capability has a small set of retrieval heads;
(2) dynamic: most of retrieval heads are activated under different contexts;
(3) intrinsic: retrieval heads are already within the base model as a consequence of large-scale pretraining. Subsequent models reuse the same set of heads. 
Our results are supported by extensive experiments on a large spectrum of models (Table~\ref{tab:all_models}).
To investigate the influence of continued pretraining for context length extension, we compare Llama-2-7B 4K to Llama-2-7B-80K and Llama-2-13B-60K \citep{fu2024data}. 
To examine the effect of alignment, we have study Mistral-7B-Instruct-v0.2 and Qwen-1.5-14B-Chat \citep{bai2023qwen} and compare them to their base versions.
We further choose Mixtral-8x7B-v0.1 \citep{jiang2024mixtral}, a mixture of expert versions derived from Mistral-7B-v0.2, presumably via sparse upcycling~\citep{komatsuzaki2022sparse}, to study retrieval heads in different architectures.

\subsection{Universal and Sparse}
Figure~\ref{fig:score_range} demonstrate that a sparse set of retrieval heads exist in all models we consider, regardless of the various pertraining, fine-tuning recipes and the underlying architecture.
Between 25\% and 52\% of the heads exhibit copy-paste behaviors at a very low frequency, with a score of between 0 and 0.1. 
Approximately 45\% to 73\% of attention heads have 0 retrieval score, meaning that they have other functionality than retrieval.
Only about 3\% to 6\% of the attention heads have a retrieval score larger than 0.1 (recall that a retrieval score 0.1 means a head retrieves at least 10\% of the tokens being asked).
It is also intriguing that although all model parameters, particularly the total number to attention heads, are largely different from each other, their ratio of retrieval heads lays in the same interval (about 5\%). 

\begin{figure*}[!t]
\small
  \includegraphics[width=\linewidth]{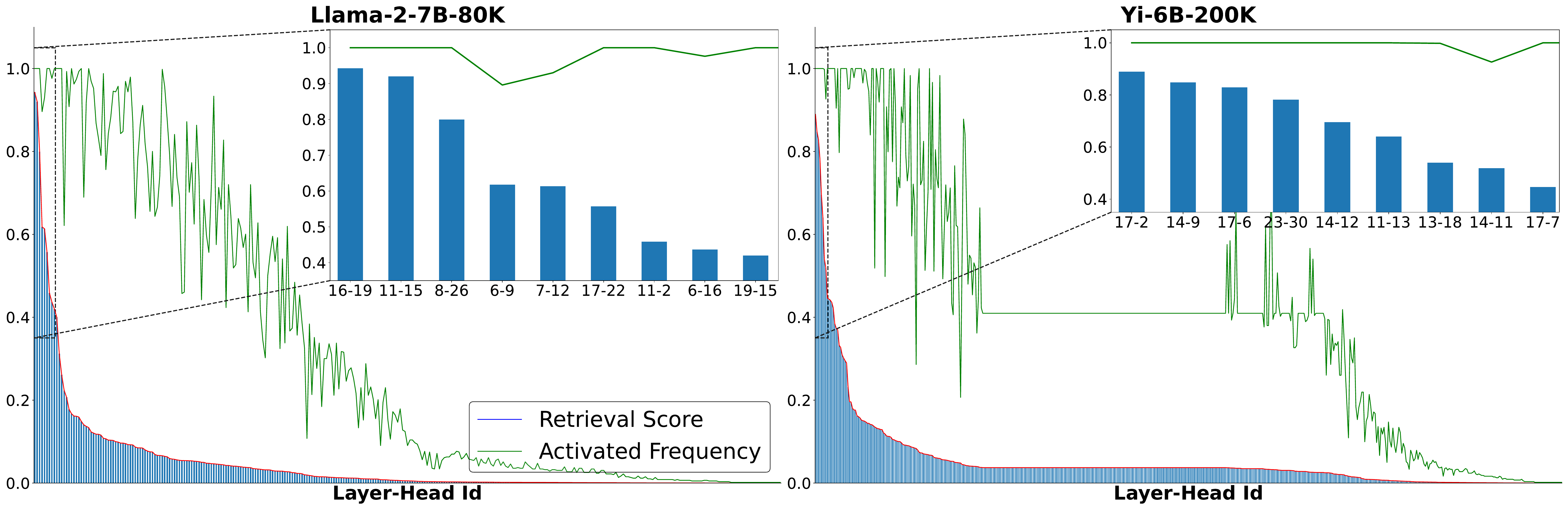}
  \caption{Retrieval score (blue): average number of activated tokens; 
  Activation Frequency (green): frequency of at least activated on one token. 
  The \textit{gap} between the two curves shows \textit{context-sensitivity}: a head of high activation frequency but low retrieval score means it is only activated on certain tokens and contexts; a head of high retrieval score means it is activated on almost any context. 
  For both LLaMA and Yi, there exist strong heads that are not sensitive to context and always activated.
  }
  \label{fig:score_dist}
\end{figure*}

\subsection{Dynamically Activated Based on Tokens and Contexts}
Now we study how sensitive a retrieval head is to its input context, i.e.,
whether a head is consistently activated no matter what the context is, or if a head is activated only on specific contexts.
For the needle sentences "the best thing to do in San Francisco is eating a sandwich in Dolores park in a sunny day", some heads are activated on the full sentence, whereas other heads  only activated on certain tokens like ``eating a sandwich'' or ``in Dolores park'. 
We define \textit{activation frequency}, the frequency of a head being activated on \textit{at least one token} (v.s., the retrieval score measures the \textit{average} number of activated tokens). 
A head of high activation frequency but low retrieval score means it is only activated on certain tokens and contexts.
As is shown in Fig.~\ref{fig:score_dist}, 
Llama-2-7B-80K and Yi-6B-200K have 12 and 36 strongest retrieval heads, respectively, that are always activated (activation frequency equal to 1) under all the contexts we consider. 
Weaker heads only activate on certain tokens and contexts.

\subsection{Intrinsic}
\begin{figure*}[t!]
\small
  \centering
  \includegraphics[width=\linewidth]{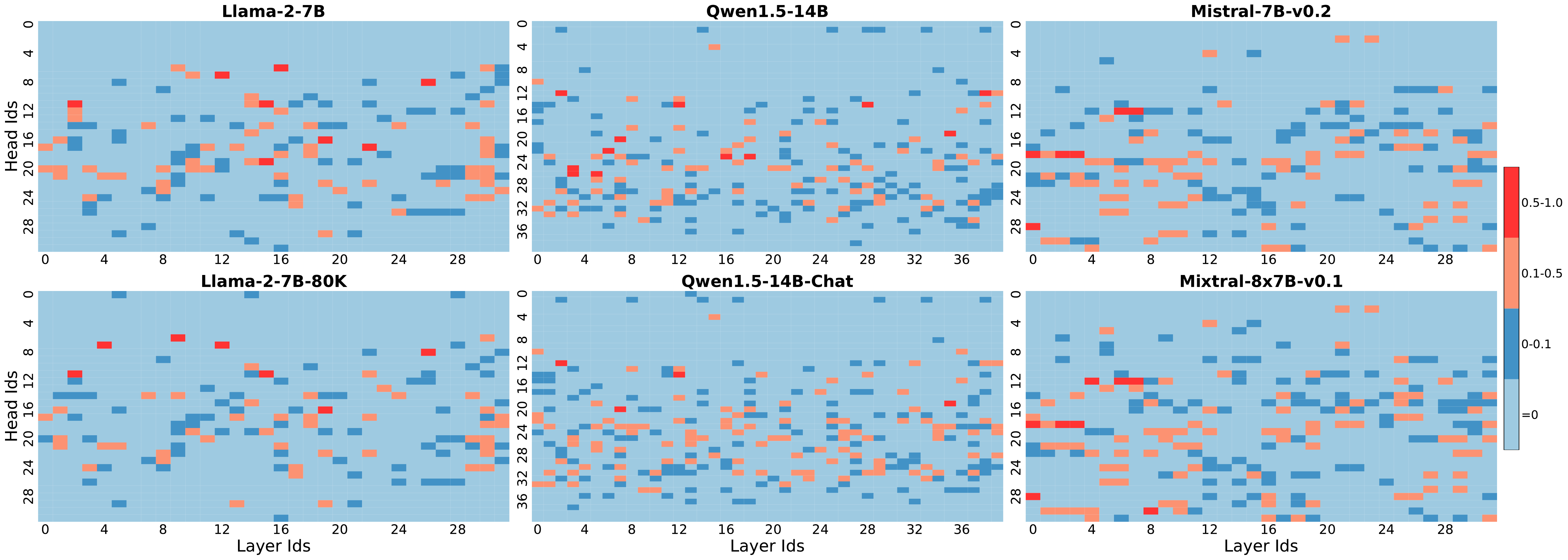}
\caption{Retrieval head is intrinsic and already within the base model. The subsequent model derivations, either by continue pretraining (LLaMA 2 7B 80K) or chat finetuning (Qwen 1.5 14B Chat) or sparse upcycling (Mixtral 8$\times$7B), use the same set of retrieval head as the base model, as demonstrated by a high level of similarity between the heatmap patterns. 
}
  \label{fig:heat_map}
\end{figure*}

\begin{figure}[!t]
    \begin{minipage}[!t]{0.48\textwidth}
        \includegraphics[width=\textwidth]{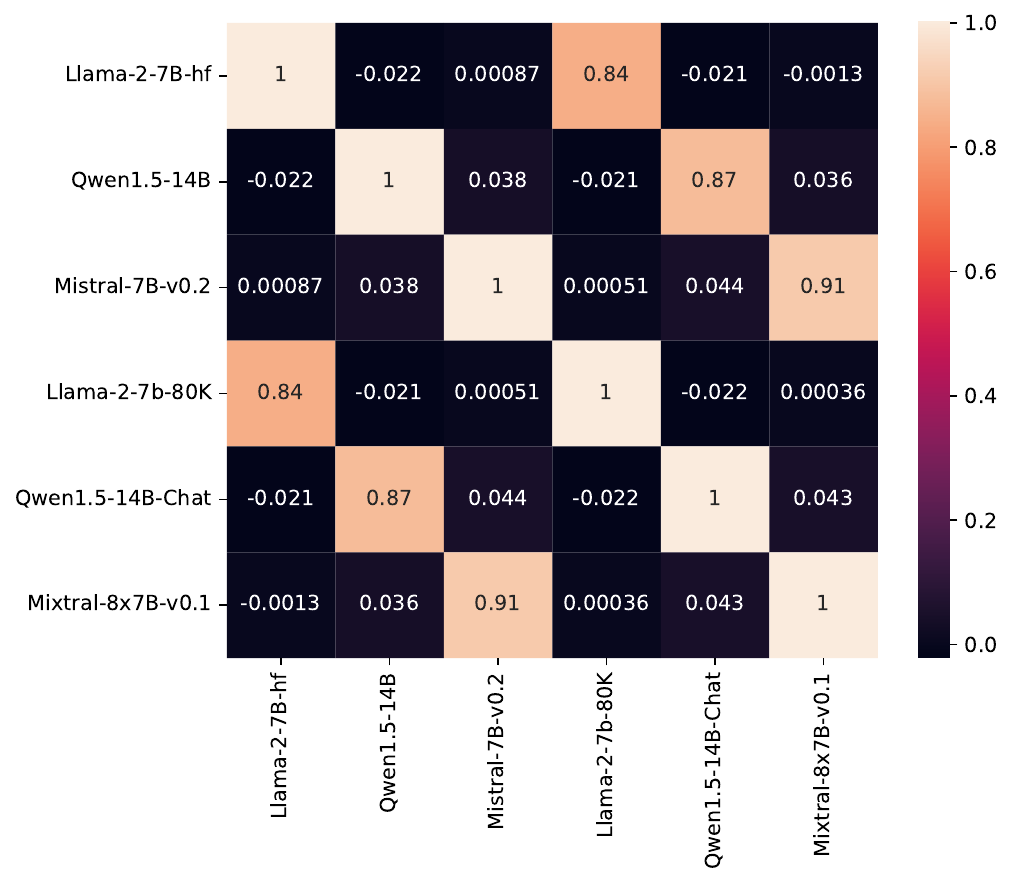}
        \caption{The retrieval heads of models of the same family are strongly correlated, i.e., the chat model and base model typically uses the same set of retrieval heads.
        The retrieval heads of models of different families are clearly different. 
        % For models of different families, their retrieval heads are clearly different. 
        }
        \label{fig:corr_map}
    \end{minipage}
    \hfill
    \begin{minipage}[t!]{0.48\textwidth}
        \includegraphics[width=\textwidth]{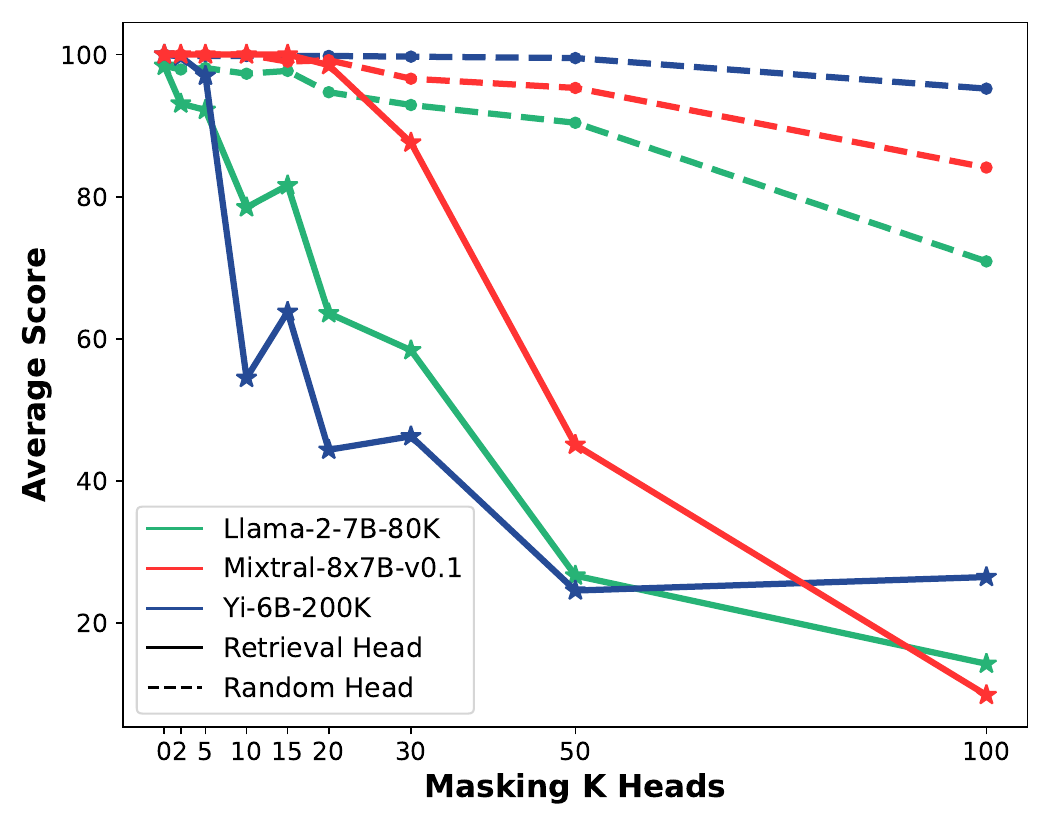}
        \caption{Masking out top K retrieval heads vs K random heads. For all models are consider, the removal of retrieval heads clearly reduces the Needle-in-a-Haystack performance, versus the removal of non-retrieval heads have much weaker influence. 
        }
        \label{fig:needle_mask}
    \end{minipage}
\end{figure}

We show that the retrieval heads, thus the ability of utilizing information at arbitrary location of the input, is an intrinsic property~\citep{fu2024data} of the base model as a consequence of large-scale pretraining, with subsequent small-scale training exerting only minor alterations to these head activation patterns.
In Figure~\ref{fig:heat_map}, we present the retrieval score distributions for a range of base models in the initial row, followed by their corresponding variants in the subsequent row. We see that regardless of the models being continuously pre-trained, chat fine-tuned, or sparsely upcycled, there is a notable consistency in their retrieval scores heatmaps. 
Figure~\ref{fig:corr_map} offers a more direct and strict examination, where we compute the statistical correlations between different models. 
The data reveal a high degree of correlation in the retrieval score distributions between base models and their respective variants, with a Pearson correlation coefficient exceeding 0.8. 
Models from different families exhibit a correlation coefficient of less than 0.1, indicative of their distinct pretraining recipes.

\section{Influence on Downstream Tasks}
\label{sec:downstream_tasks}

This section examines how retrieval heads influence downstream tasks. 
Across the experiments we use Mistrial-7B-Instruct-v0.2~\citep{mistralv02} as it is a popular and strong open language model with 32K context length. 
We first show that retrieval heads explains the factuality of Needle-in-a-Haystack test. 
When the model can retrieve the needle, retrieval heads are always activated. 
When the model cannot retrieve the needle and hallucinate instead, retrieval heads are either partially activated or not activated. 
Then we show that retrieval heads significantly influence question answering that requires extracting the information from the input, but does not strongly influence tasks where the model directly produce answers based on its internal knowledge.
We further explore how retrieval heads influence more sophisticated reasoning behaviors like chain-of-thought~\citep{DBLP:conf/nips/Wei0SBIXCLZ22}.

\subsection{Retrieval Heads Explains Factuality in Needle-in-a-Haystack}
We start with a closer look at Needle-in-a-Haystack and  
construct an additional set of needle tests that are different from the three sets used for retrieval head detection. 
We gradually mask out the number of retrieval/ random heads and see how the model's behavior changes. 
As is shown in Fig.~\ref{fig:needle_mask}, masking out retrieval heads severely damages the model's Needle-in-a-Haystack performance, while masking out random heads shows much smaller performance impact. 
Notably, when increasing the number of masked heads K to 50 (about 5\% of the full number of heads), all models' needle test performance drop to below 50, showing that the top retrieval heads are responsible for most of the needle retrieval behavior.

\begin{figure}[!t]
    \begin{minipage}[t]{0.3\textwidth}
        \vspace{0pt}
        \includegraphics[width=\textwidth]{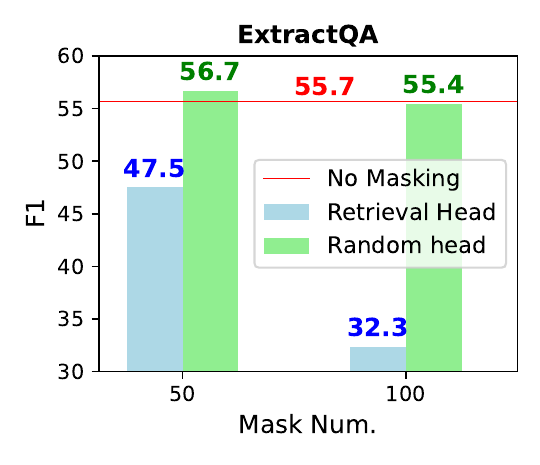}
        \caption{Masking out retrieval heads severely damages ExtractQA performance. 
        }
        \label{fig:task_qa}
    \end{minipage}
    \hfill
    \begin{minipage}[t]{0.66\textwidth}
        \vspace{0pt}
        \includegraphics[width=\textwidth]{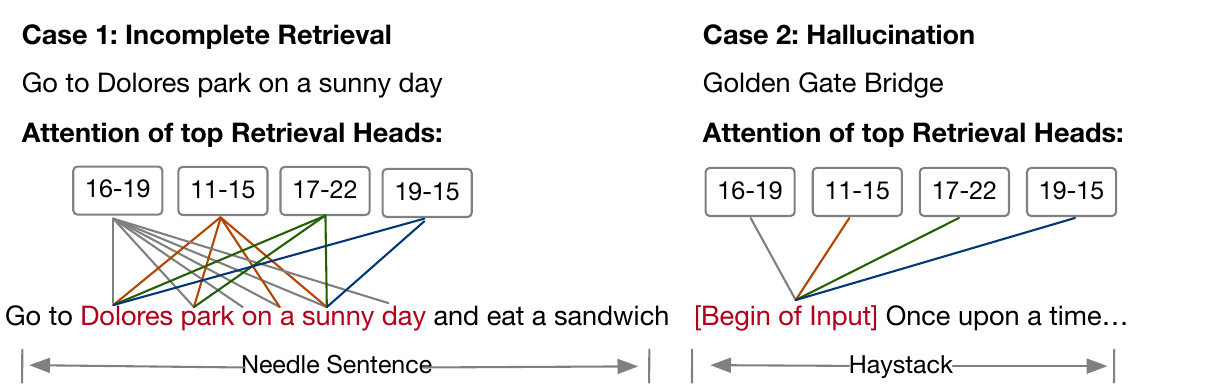}
        \caption{When the model fails to retrieve the full needle, there are two typical errors:
(1) incomplete retrieval, where the retrieval heads miss part of the information ``eat a sandwich'';
(2) hallucination, where the retrieval heads attend to the initial tokens.
        }
        \label{fig:wrong_head_case}
    \end{minipage}
\end{figure}
We observe three types of error: (1) Incomplete retrieval, where models only captured partial information of the target and omitted key details; (2) Hallucination, where models generated hallucinated sentences; and (3) Wrong extraction, where models incorrectly retrieved irrelevant content from the haystack.
Without any masking, instances of wrong extraction occurred when retrieval heads were active but focused on incorrect sections. During hallucinated generation, retrieval heads predominantly attended to the initial token of the input, which is often known as "attention sink"~\citep{xiao2023efficient}, presumably dummy tokens that provide less meaningful signal.

As we increase the number of masked heads, initially, 
a small subset of the most robust retrieval heads are masked, and incomplete retrieval began to appear. 
In the absence of the strongest retrieval heads, the remaining weaker heads only managed to retrieve a fraction of the target information.
Metaphorically, each retrieval head holds a small piece of the "needle," yet these pieces cannot form a complete one, resulting in an incomplete final output. 
This phenomenon typically begins when the mask out heads of  retrieval score larger than 0.4.
As we further increase the number of mask, 
% In the second stage, as more retrieval heads are masked, 
hallucinations become more prevalent, signaling a complete failure of the retrieval capability.

\subsection{Influence on Extractive QA}
Now we study how retrieval heads influence more realistic tasks beyond Needle-in-a-Haystack. 
We use extractive QA as a test bed, as common usecase of long-context model where the user typically upload a pdf (research papers, finantial reports, legal documents, .etc) and ask questions about specific information within the document. 
To make sure the knowledge being asked does not exist in the model's internal knowledge, we synthesize an extractive QA dataset by selecting a set of \textit{up-to-date} news articles, extract a paragraph from it, and asking GPT-4 to produce a question-answer pair based on the extracted paragraph, similar to the evaluation conducted in~\citet{claude}. 
As illustrated in Figure~\ref{fig:task_qa},
randomly masking out non-retrieval heads demonstrated no significant impact on performance. 
Masking out retrieval heads led to a substantial decrease in F1 scores, with reductions of 9.2\% and 23.1\%.
These observations demonstrate that real-world document QA tasks heavily rely on the functionality of retrieval heads.

\subsection{Chain-of-Thought Reasoning also Requires Retrieval Heads}
\begin{figure*}[t!]
\small
  \centering
  \includegraphics[width=1.0\linewidth]{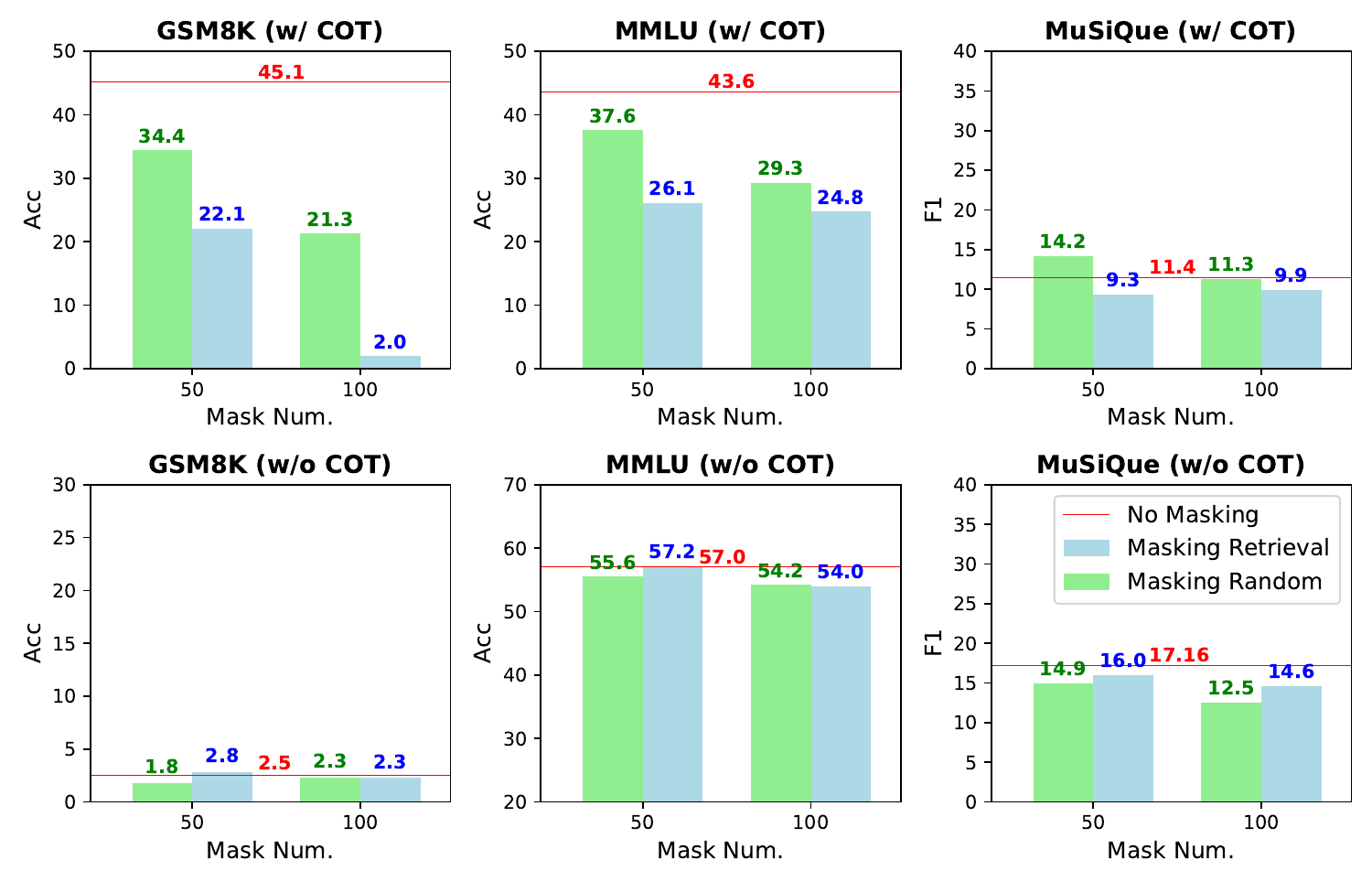}
  \caption{Retrieval heads significantly influence tasks that require chain-of-thought reasoning. 
  This is because typically in a reasoning chain, the next step reasoning requires the model to refer to previous information. See Fig.~\ref{fig:gsm8k_error_case} for examples.
  }
  \label{fig:task_cot}
\end{figure*}

\begin{figure}[t!]
\small
  \centering
  \includegraphics[width=\linewidth]{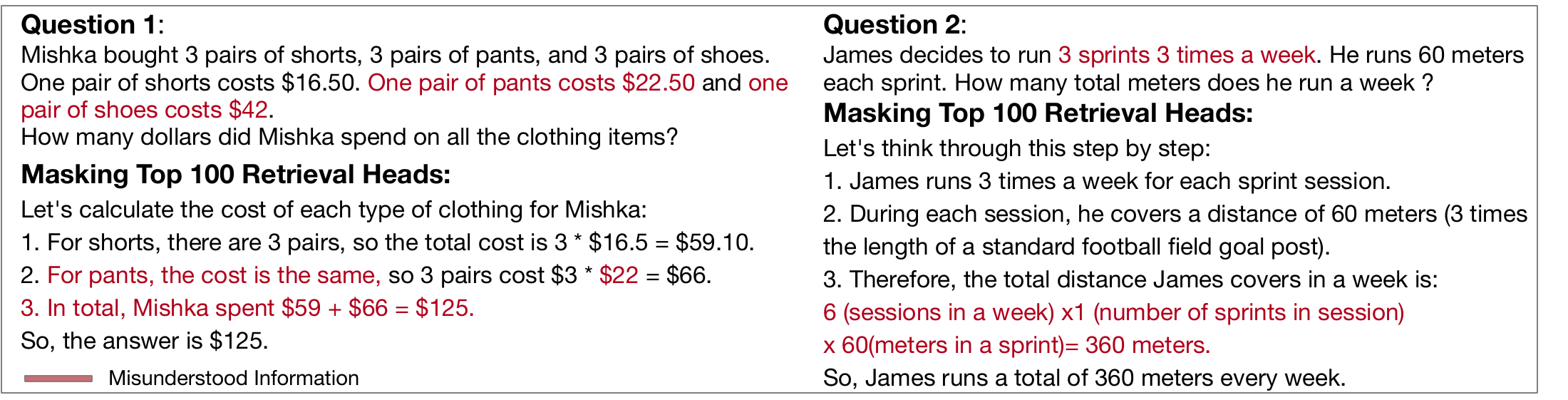}
  \caption{ 
  % Two cases of the generated CoT inferenced without retrieval heads from GSM8K.
  % Red lines are the  information  the language model failed to use, and its corresponded generation.  Blue lines are irrelevant  hallucinations
  % When masking out retrieval heads, 
  When we mask out retrieval heads, the model becomes ``blind'' to important information in the question description
  % the model fails to utilize important input information and hallucinates instead,
  resulting in incorrect reasoning chains. 
  }
  \label{fig:gsm8k_error_case}
\end{figure}
We test Mistrial-7B-Instruct-v0.2's performance on MMLU~\citep{hendrycks2020measuring}, MuSiQue and GSM8K~\citep{cobbe2021training}, with and without chain-of-thought reasoning. 
As is shown in Fig.~\ref{fig:task_cot}, 
if we use answer-only prompting (without CoT), masking out either retrieval or random heads do not really influence the performance, presumably because the model's generation is based on its internal knowledge primarily stored in the FFN layers~\citep{geva2020transformer}. 
For CoT styled reasoning, masking out retrieval heads signifantly influence the model's performance. 
Upon inspecting typically error cases (Fig.~\ref{fig:gsm8k_error_case}), we find out the model becomes ``blind'' to important input information and hallucinate instead. 
We find the relationship between CoT and retrieval heads particularly intriguing as it may offers deeper insights into model's complex reasoning performance. 
We leave more in-depth studies to future research.

% \section{The Super Compression Hypothesis}
% \label{sec:discussions}
% \input{040_discussion}

\section{Discussions}
\label{sec:discussions}
\textbf{General Functionalities of Attention Heads}\quad\quad
For transformer language models, we tend to view the functionality of FNNs layers to be the place for storing knowledge~\citep{geva2020transformer}, and the attention layers to be the place for implementing algorithms~\citep{olsson2022context}. 
The induction head discussed in ~\citet{olsson2022context} typically searches repeated patterns of the input, which is at a certain level similar to the retrieval heads (as it also searches and repeats information). 
Different than the induction heads, the retrieval heads are typically responsible for redirecting the information according to the context, but do not for inferring programs. 
We tend to believe that there exist more algorithm and functionalities implemented by other types of attention heads to be discovered by future research.

\textbf{Relationship to Local and Linear Attention and State-Space Models}\quad\quad
Although there exist numerous works about local~\citep{xiao2023efficient} / linear~\citep{wang2020linformer} attention, state space models~\citep{gu2023mamba}, and hybrid architectures~\citep{de2024griffin} achieving inspiring efficiency in long-context modeling, so far there is no linear attention / SSM architecture that passes the Needle-in-a-Haystack test to the best of our knowledge, suggesting that the full attention might be a must for long-context information retrieval. 
One example is that the Mistral v0.1~\citep{jiang2023mistral} uses sliding window attention but cannot pass needle-in-a-haystack, and their authors changes the attention to full in v0.2~\citep{mistralv02}, then it can pass the needle test. 
Our results showing strong evidence why full attention is a must.  
For the model to precisely utilize input information at arbitrary location, it is crutial for the retrieval heads to work on the full KV cache.

\textbf{Applications to KV Cache Compression}\quad\quad
The problem that the KV cache is too large and occupies a large chunk of the GPU memory severely hinders the deployment of long-context models.
For example, for LLaMA 2 7B, the KV cache of 100K tokens requires more than 50GB memory, while 2K context requires less than 1GB memory. 
If we serve this model on one 80G A100, then the concurrency of 100K context can be 50 times less than 2K context queries, which is prohibitively expensive. 
The results from this work indicates that we might be possible to radically prune out the KV cache corresponding to the non-retrieval heads (recall in Fig.~\ref{fig:score_range} shows only 5\% of the heads are retrieval) and significantly reducing the deployment cost of long-context models. 
We leave this study to future research.
% typically we would say this is a future research, but it is done concurrently by Xiao et. al. 2024

\section{Conclusions}
\label{sec:conclusions}
This paper discovers retrieval heads, a special set of attention heads that are responsible for implementing the conditional copy algorithm and redirect information from the input to the output. 
Retrieval heads are the primarily reason why a successful long-context model can pass the Needle-in-a-Haystack test, and their activation explains why a language model is faithful to the input or hallucinate.
Compared to non-retrieval heads, retrieval heads have a stronger influence on downstream tasks that require the model to precisely recall the input information, either in extractive question answering or chain-of-thought reasoning. 
We believe this work will foster future research on reducing hallucination,
improving reasoning, and compressing the KV cache. 

\bibliographystyle{plainnat}
\bibliography{retrieval_head}

%%%%%%%%%%%%%%%%%%%%%%%%%%%%%%%%%%%%%%%%%%%%%%%%%%%%%%%%%%%%

\end{document}